%% file: main.tex
\documentclass[11pt,a4paper]{article}
\usepackage[hyperref]{emnlp2020}
\usepackage{times}
\usepackage{latexsym}
\aclfinalcopy

\usepackage{todonotes}
\usepackage{comment}

\usepackage[T1]{fontenc}
\usepackage[utf8]{inputenc}
\usepackage[bottom]{footmisc}
\usepackage{newtxtext,newtxmath}

\usepackage{booktabs} % For formal tables
\usepackage{float}
\usepackage{silence}
\usepackage{ctable}
\usepackage{tabularx}
\usepackage{etoolbox}
\usepackage{MnSymbol}
\usepackage{hyperref}
\usepackage{amsmath} % theoremstyls
\usepackage{listings}
\usepackage{fancybox}
\usepackage{wrapfig}
\usepackage{graphicx}
\usepackage{url}
\usepackage{paralist}
\usepackage{multirow}
\usepackage[flushleft]{threeparttable}
\usepackage{subfig}

\usepackage{color, colortbl}
\definecolor{Gray}{gray}{0.9}
\definecolor{Cyan}{RGB}{204, 255, 255}
\definecolor{Yellow}{RGB}{255, 242, 204}
\definecolor{Red}{RGB}{255, 204, 204}
\newcolumntype{a}{>{\columncolor{Gray}} c}

\usepackage{pgf}
\usepackage{ifthen} 
\usepackage{dcolumn}
\newcolumntype{d}[1]{D{.}{.}{#1}}
\newcolumntype{L}[1]{>{\raggedright\let\newline\\\arraybackslash\hspace{0pt}}m{#1}}

\graphicspath{{figures/}}
\DeclareGraphicsExtensions{.pdf}

\graphicspath{{figures/}}
\DeclareGraphicsExtensions{.pdf}

\newenvironment{nospaceflalign*}
 {\setlength{\abovedisplayskip}{-3pt}\setlength{\belowdisplayskip}{4pt}%
  \csname flalign*\endcsname}
 {\csname endflalign*\endcsname\ignorespacesafterend}

\newcommand{\myurl}[1]{{\fontsize{7}{8}\selectfont{\url{#1}}}}

\newcommand{\myequationfont}{
\fontsize{8}{9}\selectfont
}
\newcommand{\mycodefont}{
\fontsize{8}{9}\selectfont\ttfamily
}

\newcommand{\code}[1]{\texttt{#1}}
\usepackage{pifont}% http://ctan.org/pkg/pifont

\newsavebox{\measurebox}

\newcommand{\question}[1]{\emph{\qq{#1}}}
\newcommand{\ie}{i.e.~}

\newcommand{\eg}{e.g.~}

\newcommand{\qq}[1]{``#1''}

\usepackage{marginnote}
\usepackage{todonotes}

\usepackage{microtype}

\title{QA2Explanation: Generating and Evaluating Explanations for Question Answering Systems over Knowledge Graph
}

\author{Saeedeh Shekarpour \\
  University of Dayton \\
  Dayton, USA \\
  \texttt{sshekarpour1}\\
  \texttt{@udayton.edu} \\\And
  Abhishek Nadgeri\\
  RWTH Aachen \& \\
Zerotha Research, Germany\\
\texttt{abhishek.nadgeri} \\
\texttt{@rwth-aachen.de} \\\And
  Kuldeep Singh \\
  Cerence GmbH \&\\
  Zerotha Research, Germany  \\
  \texttt{kuldeep.singh1} \\
  \texttt{@cerence.com} \\}

\date{2020}

\begin{document}
\maketitle
\begin{abstract}
In the era of Big Knowledge Graphs, Question Answering (QA) systems have reached a milestone in their performance and feasibility.
However, their applicability, particularly in specific domains such as the biomedical domain, has not gained wide acceptance due to their ``black box'' nature, which hinders transparency, fairness, and accountability of QA systems. Therefore, users are unable to understand how and why particular questions have been answered, whereas some others fail.
To address this challenge, in this paper, we develop an automatic approach for generating explanations during various stages of a pipeline-based QA system. 
Our approach is a supervised and automatic approach which considers three classes (i.e., success, no answer, and wrong answer) for annotating the output of involved QA components.
Upon our prediction, a template explanation is chosen and integrated into the output of the corresponding component.
To measure the effectiveness of the approach, we conducted a user survey as to how non-expert users perceive our generated explanations. The results of our study show a significant increase in the four dimensions of the human factor from the Human-computer interaction community. 
\end{abstract}

\input{body}

\bibliographystyle{acl_natbib}
\bibliography{ref}

\end{document}

%% file: body.tex
\section{Introduction}
%%%%%%%%%%%%%%%%%%%%%%%%%%%%%%
The recent advances of Question Answering (QA) technologies mostly rely on (i) the advantages of Big Knowledge Graphs which augment the semantics, structure, and accessibility of data, \textit{e.g.,} Web of Data has published around 150B triples from a variety of domains\footnote{\url{http://lodstats.aksw.org/}}, and (ii) the competency of contemporary AI approaches which train sophisticated learning models (statistical models \cite{sina,shekarpour:www2013}, neural networks \cite{www/LukovnikovFLA17}, and attention models \cite{liu2019conditioning}) on a large size of training data, and given a variety of novel features captured from semantics, structure, and context of the background data.
However, similar to other branches of AI applications, the state of the art of QA systems are \textit{``black boxes''} that fail to provide transparent explanations about why a particular answer is generated. This black box behavior diminishes the confidence and trust of the user and hinders the reliance and acceptance of the black-box systems, especially in critical domains such as healthcare, biomedical, life-science, and self-driving cars \cite{samek2017explainable,miller2018explanation}.
%The black box behavior diminishes the confidence and trust of the user in wider adaptability of various systems such as expert systems, intelligent systems, and recommender systems in real world scenario \cite{miller2018explanation}. 
The \underline{running hypothesis} in this paper is that the lack of explanation for answers provided by  QA systems diminishes the trust and acceptance of the user towards these systems. 
Therefore, by implementing more transparent, interpretable, or explainable QA systems, the end users will be better equipped to justify and therefore trust the output of QA systems \cite{li2018tell}. 

Furthermore, \textbf{data quality} is a critical factor that highly affects the performance of QA systems. In other words, when the background data is flawed or outdated, it undermines the human-likeness and acceptance of the QA systems if no explanation is provided, especially for non-expert users.
For example, the SINA engine \cite{sina} failed to answer the simple question \question{What is the population of Canada?} on the DBpedia \cite{DBLP:conf/semweb/AuerBKLCI07} version 2013, whereas it succeeded for similar questions such as \question{What is the population of Germany?}. The error analysis showed that the expected triple \textit{i.e.,} <\texttt{dbr\footnote{\texttt{dbr} is bound to \url{http://dbpedia.org/resource/}.}:Canada dbo\footnote{The prefix \code{dbo}\ is bound to \url{http://dbpedia.org/ontology/}.}:population "xxx"}> is missing from DBpedia 2013. Thus, if the QA system does not provide any explanation about such failures, then the non-expert user concludes the QA system into the demerit points.
Thus, in general, the errors or failures of the QA systems might be caused by the inadequacies of the underlying data or misunderstanding, misinterpretation, or miscomputation of the employed computational models.
In either case, the black-box QA system does not provide any explanations regarding the sources of the error. 
Often the research community obsesses with the technical discussion of QA systems and competes on enhancing the performance of the QA systems, whereas, on the downside of the QA systems, there is a human who plays a vital role in the acceptance of the system. 
The Human-Computer Interaction (HCI) community already targeted various aspects of the human-centered design and evaluation challenges of black-box systems. However, the QA systems over KGs received the least attention comparing to other AI applications such as recommender systems \cite{herlocker2000explaining,kouki2017user}.

\textbf{Motivation and Approach:} 
Plethora of QA systems over knowledge graphs developed in the last decade \cite{DBLP:journals/semweb/HoffnerWMULN17}. These QA systems are evaluated on various benchmarking datasets including WebQuestions \cite{DBLP:conf/emnlp/BerantCFL13}, QALD \cite{DBLP:conf/clef/UngerFLNCCW15}, LC-QuAD \cite{trivedi2017lc}, and report results based on global metrics of precision, recall, and F-score. In many cases, QA approaches over KGs even surpass the human level performance \cite{petrochuk2018simplequestions}. Irrespective of the underlying technology and algorithms, these QA systems act as black box and do not provide any explanation to the user regarding 1) why a particular answer is generated and 2) how the given answer is extracted from the knowledge source. The recent works towards explainable artificial intelligence (XAI) gained momentum because several AI applications find limited acceptance due to ethical reasons \cite{pro} and a lack of trust on behalf of their users \cite{stubbs2007autonomy}. The same rationale is also applicable to the black-box QA systems. Research studies showed that representing adequate explanations to the answer brings acceptability and confidence to the user as observed in various domains such as recommender systems and visual question answering \cite{herlocker2000explaining,hayes2017improving,hendricks2016generating,wu2018faithful}. 
In this paper, we argue that having explanations increases the trustworthiness, transparency, and acceptance of the answers of the QA system over KGs. Especially, when the QA systems fail to answer a question or provide a wrong answer, the explanatory output helps to keep the user informed about a particular behavior. Hence, we propose a template-based explanation generation approach for QA systems. Our proposed approach for explainable QA system over KG provides (i) \textit{adequate justification:} thus the end user feels that they are aware of the reasoning steps of the computational model, (ii) \textit{confidence:} the user can trust the system and has the willing for the continuation of interactions, (iii) \textit{understandability:} educates the user as how the system infers or what are the causes of failures and unexpected answers, and (iv) \textit{user involvement:} encourages the user to engage in the process of QA such as question rewriting.

\textbf{Research Questions:} We deal with two key research questions about the explanations of the QA systems as follows: \textbf{RQ1:} \emph{What is an effective model and scheme for automatically generating explanations?}
The computational model employed in a QA system might be extremely complicated.
The exposure of the depth of details will not be sufficient for the end user. The preference is to generate natural language explanations that are readable and understandable to the non-expert user.
\textbf{RQ2:} \emph{How is the perception of end users about explanations along the human factor dimensions?}, which is whether or not the explanations establish confidence, justification, understanding, and further engagements of the user.

Our key contributions are: 1) a scheme for shallow explanatory QA pipeline systems, 2) a method for automatically generating explanations, and 3) a user survey to measure the human factors of user perception from explanations. 
This paper is organized as follows: In Section \ref{sec:related-work}, we review the related work.
Section \ref{sec:pipeline} explains the major concepts of the QA pipeline system, which is our employed platform. Section \ref{sec:approach} provides our presentation and detailed discussion of the proposed approach. 
Our experimental study is presented in Section \ref{sec:experiment}, followed by a discussion Section.
We conclude the paper in section \ref{sec:conclusion}.
%%%%%%%%%%%%%%%%%%%%%%%%%%%%%%%%%%%%%%%%%%%%%%%%%%%%%%%%%%%%%%%%%%%%%%%%%%%%%%%%%%%%%%%%%%%%%%%%%%%%%%%%%%%
%%%%%%%%%%%%%%%%%%%%%%%%%%%%%%%%%%%%%%%%%%%%%%%%%%%%%%%%%%%%%%%%%%%%%%%%%%%%%%%%%%%%%%%%%%%%%%%%%%%%%%%%%%%
%%%%%%%%%%%%%%%%%%%%%%%%%%%%%%%%%%%%%%%%%%%%%%%%%%%%%%%%%%%%%%%%%%%%%%%%%%%%%%%%%%%%%%%%%%%%%%%%%%%%%%%%%%%
%%%%%%%%%%%%%%%%%%%%%%%%%%%%%%%%%%%%%%%%%%%%%%%%%%%%%%%%%%%%%%%%%%%%%%%%%%%%%%%%%%%%%%%%%%%%%%%%%%%%%%%%%%%
%%%%%%%%%%%%%%%%%%%%%%%%%%%%%%%%%%%%%%%%%%%%%%%%%%%%%%%%%%%%%%%%%%%%%%%%%%%%%%%%%%%%%%%%%%%%%%%%%%%%%%%%%%%
%%%%%%%%%%%%%%%%%%%%%%%%%%%%%%%%%%%%%%%%%%%%%%%%%%%%%%%%%%%%%%%%%%%%%%%%%%%%%%%%%%%%%%%%%%%%%%%%%%%%%%%%%%%
%%%%%%%%%%%%%%%%%%%%%%%%%%%%%%%%%%%%%%%%%%%%%%%%%%%%%%%%%%%%%%%%%%%%%%%%%%%%%%%%%%%%%%%%%%%%%%%%%%%%%%%%%%%
%%%%%%%%%%%%%%%%%%%%%%%%%%%%%%%%%%%%%%%%%%%%%%%%%%%%%%%%%%%%%%%%%%%%%%%%%%%%%%%%%%%%%%%%%%%%%%%%%%%%%%%%%%%
\section{Related Work}
\label{sec:related-work}
Researchers have tackled the problem of question answering in various domains including open domain question answering \cite{yang2019end}, biomedical \cite{bhandwaldar2018uncc}, geospatial \cite{punjani2018template}, and temporal \cite{jia2018tequila}. Question answering over publicly available KGs is a long-standing field with over 62 QA systems developed since 2010 \cite{DBLP:journals/semweb/HoffnerWMULN17}. The implementation of various QA systems can be broadly categorized into three approaches \cite{singh2019towards,diefenbach2017core}. The first is a semantic parsing based approach such as \cite{usbeck2015hawk} that implements a QA system using several linguistic analyses (e.g., POS tagging, dependency parsing) and linked data technologies. 
The second approach is an end-to-end machine learning based, which uses a large amount of training data to map an input question to its answer directly (e.g., in \cite{yang2019end,www/LukovnikovFLA17}). The third approach is based on modular frameworks \cite{kimokbqa,www/SinghRBSLUVKP0V18} which aims at reusing individual modules of QA systems, independent tools (such as entity linking, predicate linking) in building QA systems collaboratively. Irrespective of the implementation approach, domain, and the underlying knowledge source (KG, documents, relational tables, etc.), the majority of existing QA systems act as a black box. The reason behind black box behavior is due to either the monolithic tightly coupled modules such as in semantic parsing based QA systems or nested and nonlinear structure of machine learning based algorithms employed in QA systems. The modular framework, on the other hand, provides flexibility to track individual stages of the answer generation process. The rationale behind our choice of the modular framework over monolithic QA systems is a flexible architecture design of such frameworks. It allows us to trace failure at each stage of the QA pipeline. We enrich the output of each step with adequate justification with supporting natural language explanation for the user. Hence, as the first step towards explainable QA over knowledge graphs, we propose an automatic approach for generating a description for each stage of a QA pipeline in a state-of-the-art modular framework (in our case: Frankenstein \cite{www/SinghRBSLUVKP0V18}).  
We are not aware of any work in the direction of explainable question answering over knowledge graphs and we make the first attempt in this paper. Although, efforts have been made to explain visual question answering systems. Some works generate textual explanations for VQA by training a recurrent neural network (RNN) to mimic
examples of human descriptions \cite{hendricks2016generating,wu2018faithful} directly. The work by \cite{ngonga2013sorry} can be considered a closest attempt to our work. The authors proposed a template based approach to translate SPARQL queries into natural language verbalization. We employ a similar template-based approach to generate an automatic explanation for QA pipelines.

In other domains, such as expert systems, the earlier attempts providing explanations to the users can be traced back in the early 70s \cite{shortliffe1974rule}. Since then, extensive work has been done to include explanations in expert systems followed by recommender systems to explain the system's knowledge of the domain and the reasoning processes these systems employ to produce results (for details, please refer to \cite{moore1988explanation,jannach2010recommender,daher2017review}. For a recommender system, work by \cite{herlocker2000explaining} is an early attempt to evaluate different implementations of explanation interfaces in "MovieLens" recommender system. Simple statements provided to the customers as explanations mentioning the similarity to other highly rated films or a favorite actor or actress were among the best recommendations of the MovieLens system compared to the unexplained recommendations. Furthermore, applications of explanation are also considered in various sub-domains of artificial intelligence, such as justifying medical decision-making \cite{fox2007argumentation}, explaining autonomous agent behavior \cite{hayes2017improving}, debugging of machine learning models \cite{kulesza2015principles}, and explaining predictions of classifiers \cite{ribeiro2016should}.

%recommendation systems, the earlier attempts to provide explanation can be traced back in as early as year 2000.

%%%%%%%%%%%%%%%%%%%%%%%%%%%%%%%%%%%%%%%%%%%%%%%%%%%%%%%%%%%%%%%%%%%%%%%%%%%%%%%%%%%%%%%%%%%%%%%%%%%%%%%
\section{QA Pipeline on Knowledge Graph} \label{sec:pipeline}
One of the implementation approaches for answering questions from interlinked knowledge graphs is typically a multi-stage process which is called \textit{QA pipeline} \cite{www/SinghRBSLUVKP0V18}.
Each stage of the pipeline deals with a required task such as Named Entity Recognition (NER) and Disambiguation (NED) (referred as Entity Linking (EL)), Relation extraction and Linking (RL), and Query Building (QB).
There is an abundance of components performing QA tasks~\cite{diefenbach2017core}. 
These implementations run on the KGs and have been developed based on AI, NLP, and Semantic Technologies, which accomplish one or more tasks of a QA pipeline \cite{DBLP:journals/semweb/HoffnerWMULN17}.
Table~\ref{tab:motivating} \cite{www/SinghRBSLUVKP0V18} presents performance of best QA components on the LC-QuAD dataset, implementing QA tasks. The components are Tag Me API~\cite{DBLP:conf/cikm/FerraginaS10}) for NED, 
RL (Relation Linking) implemented by RNLIWOD\footnote{Component is similar to Relation Linker of \url{https://github.com/dice-group/NLIWOD}} and SPARQL query builder by NLIWOD QB\footnote{Component is based on \url{https://github.com/dice-group/NLIWOD} and \cite{DBLP:conf/www/UngerBLNGC12}.}). 
For example, given the question \question{Did Tesla win a nobel prize in physics?}, the ideal NED component is expected to recognize the keyword \question{Tesla} as a named entity and map it to the corresponding DBpedia resource, \ie \texttt{dbr:Nikola\_Tesla}.
Similarly, the multi-word unit \question{nobel prize in physics} has to be linked to \texttt{dbr:Nobel\_Prize\_in\_Physics}. Thereafter, a component performing RL finds embedded relations in the given question and links them to appropriate relations of the underlying knowledge graph.
In our example, the keyword \question{win}\ is mapped to the relation \texttt{dbo:award}. 
Finally, the QB component generates a formal query (\eg expressed in SPARQL) (\ie \texttt{ASK \{dbr:Nikola\_Tesla dbo:award dbr:Nobel\_Prize\_in\_Physics.\}}). The performance values in \autoref{tab:motivating} are averaged over the entire query inventory.

\begin{table}[hb!]
	\centering
	\caption{Performance of QA components implementing various QA tasks on LC-QuAD dataset.}
	\resizebox{0.96\columnwidth}{!}{%
        \begin{tabular}{ l l l l l }
    	    \toprule
            \textbf{QA Component} & \textbf{QA Task} & \textbf{Precision} & \textbf{Recall} & \textbf{F-Score} \\
            \midrule
            {\it TagMe}
                & NED & 0.69 & 0.66 & 0.67 \\
            {\it RNLIWOD}
                & RL & 0.25 & 0.22 & 0.23 \\
            {\it NLIWOD QB}
                & QB & 0.48 & 0.49 & 0.48 \\
            \bottomrule
        \end{tabular}
        }
    \label{tab:motivating}
\end{table} 

%#######################################################

\begin{figure*}[hpbt]
\centering
    \includegraphics[width=1.8\columnwidth]{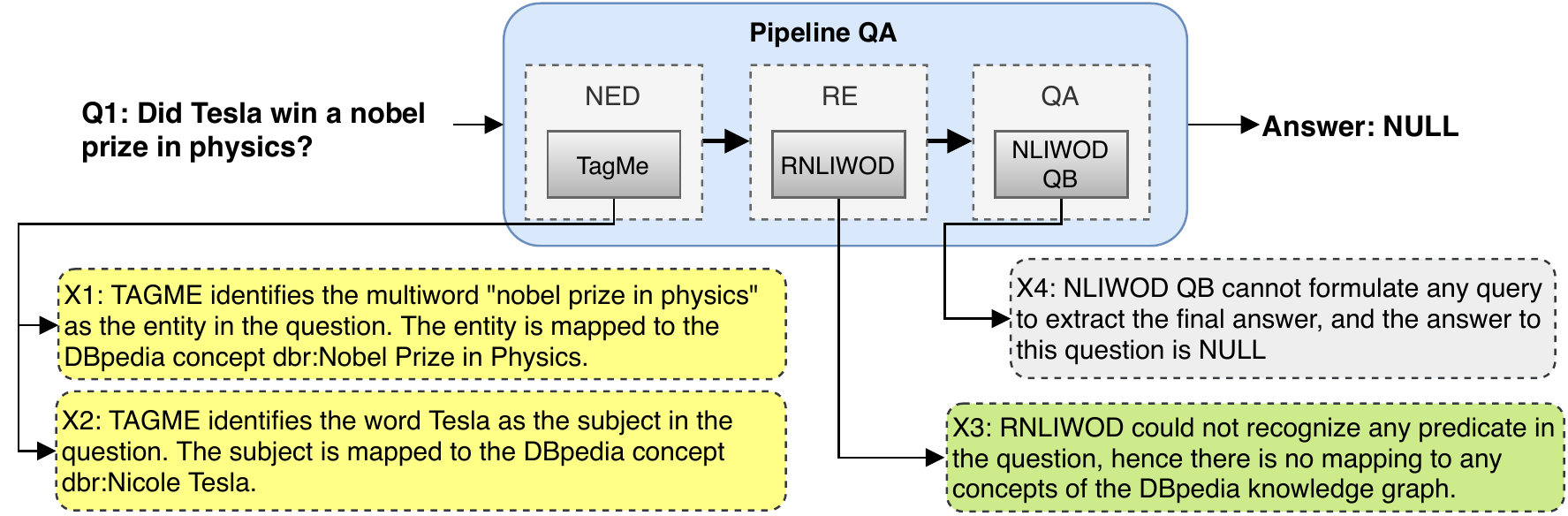}
    \caption{The QA pipeline generates the explanations in various stages of running; each explanation is generated per output of each integrated component. The demonstrated pipeline contains three components, i.e., NED, RL, and QB; the output(s) of each one is integrated into an explanation template and represented to the end user.}
     \label{fig:example}
    \end{figure*}

%########################################

%##################
\section{Approach}\label{sec:approach}
A full QA pipeline is required to answer a given question $q$. Such QA pipelines are composed of all the required components performing necessary tasks to transform a user-supplied natural language (NL) question into a formal query language (\textit{i.e.,} SPARQL). 
We consider three generic classes for outputs of a full QA pipeline or individual components, namely $O_c=\{Success, No Answer, Wrong Answer\}$.
Concerning a given question, a ``success'' class is when the QA pipeline (component) successfully provides a correct output, a ``No Answer'' class happens when the full QA pipeline (or an individual component) does not return any output and ``Wrong Answer'' class is when the provided output is incorrect.
%Figure \ref{fig:pipeline} shows a black-box QA system, which does not provide any explanation neither for a success case (Q1) nor for a failure case (Q2). 

To address \textbf{RQ1}, we introduce a scheme for generating explanations for the QA pipeline system. 
This scheme produces shallow, however automatic explanations using a semi-supervised approach for generating individual explanations after running each integrated component. 
In our proposed model, the class of the output of each integrated component is predicted using a supervised learning approach. 
We train a classifier per component within the pipeline. Then based on the prediction of the classifier, an explanation template is chosen. The explanation template and the output of the component are incorporated to form the final representation of explanations. We have a repository of explanation templates for each component of the QA pipeline system. For example, the NED component corresponds to several explanation templates differing based on the number of the output entities. Precisely, the explanation template when the NED has one single entity is different from when it has two or three. Moreover, the templates vary based on the Part of Speach (POS) tag of the entities recognized in the input question. 
For example, Figure \ref{fig:example} shows a pipeline containing three components: 1) NED component: TagMe, 2) RL component: RNLIWOD QB, and 3) QB component: NLIWOD QB. Three classifiers were individually trained for each component. 
In this example, for the given question \question{Did Tesla win a nobel prize in physics?} the classifiers predicted the class of ``Success'' for NED and the class "No Answer" for RL and QB components. 
Thus, the explanation templates corresponding to the class of ``success'' for NED, and "No Answer" for RL and QB are filtered. Then since the NED component has two outputs, therefore, two explanations were generated for NED, whereas the remaining components show one explanation.

%#######################################################
\subsection{Predicting Output of Components}
The set of necessary QA tasks formalized as $\mathcal{T}=\{t_1,t_2,\dots,t_n\}$ such as NED, RL, and QB.
Each task ($t_i:q^* \rightarrow q^+ $) transforms a given representation $q^*$ of a question $q$ into another representation $q^+$.
For example, NED and RL tasks transform the input representation \question{What is the capital of Finland?}\ into the representation \question{What is the \code{dbo:capital} of \code{dbr:Finland}?}.
The entire set of QA components is denoted by $\mathcal{C}=\{C_1,C_2,\dots,C_m\}$. 
Each component $C_j$ solves one single QA task; $C_j^{t_i}$ corresponds to the QA task $t_i$ in $\mathcal{T}$ implemented by $C_j$.
For example, RNLIWOD implements the relation linking QA task, \ie $\mathit{RNLIWOD}^\mathit{RL}$. 
Let $\rho(C_j)$ denote the performance of a QA component, then our key objective is to predict the likelihood of $\rho(C_j)$ for a given representation $q^*$ of $q$, a task $t_i$, and an underlying knowledge graph $\lambda$. 
This is denoted as $\mathit{Pr}(\rho(C_j)|q^*,t_i,\lambda)$.
In this work, we assume a single knowledge graph (\ie DBpedia); thus, $\lambda$ is considered a constant parameter that does not impact the likelihood leading to:
\begin{equation}\label{eq:eque1}
\mathit{Pr}(\rho(C_j)|q^*,t_i) = \mathit{Pr}(\rho(C_j)|q^*,t_i,\lambda)  
\end{equation}

%Moreover, for each individual task $t_i$ and question representation $q^*$, we predict the performance of all pertaining components. 
%In other words, for a given task $t_i$, the set of components that can accomplish $t_i$ is $\mathcal{C}^{t_i} =\{C_j,\dots,C_k\}$.
%Thus, we factorise $t_i$ as follows:
%\begin{equation}\label{eq:eque2}
%\forall C_j \in \mathcal{C}^{t_i},  [\mathit{Pr}(\rho(C_j)|q^*) = \mathit{Pr}(\rho(C_j)|q^*,t_i)]
%\end{equation}
Further, we assume that the given representation $q^*$ is equal to the initial input representation $q$ for all the QA components, \ie  $q^*=q$.

\paragraph{Solution} Suppose we are given a set of NL questions $\mathcal{Q}$ with the detailed results of performance for each component per task.
We can then model the prediction goal $\mathit{Pr}(\rho(C_j)|q,t_i)$ as a supervised learning problem on a training set,  
\ie a set of questions $\mathcal{Q}$ and a set of labels $\mathcal{L}$ representing the performance of $C_j$ for a question $q$ and a task $t_i$. 
In other words, for each individual task $t_i$ and component $C_j$, the purpose is to train a supervised model that predicts the performance of the given component $C_j$ for a given question $q$ and task $t_i$ leveraging the training set.
If $|\mathcal{T}|=n$ and each task is performed by $m$ components, and the QA pipeline integrates all the $n \times m$ components, then $n \times m$ individual learning models have to be built up.\\
\textbf{Question Features.}
 Since the input question $q$ has a textual representation, it is necessary to automatically extract suitable features, \ie $\mathcal{F}(q)=(f_1,\dots,f_r)$.
In order to obtain an abstract and concrete representation of NL questions, we reused question features proposed by \cite{www/SinghRBSLUVKP0V18,qaldgen2019} which impact the performance of the QA systems. These features are: question length, answer type (list, number, boolean), Wh-word (who,what,which,etc.), and POS tags present in a question. Please note, our contribution is not the underlying Frankenstein framework, we reused it for the completion of the approach. Our contribution is to add valid explanation to each step of the QA pipeline, and empirical study to support our hypothesis.
\begin{figure}[hpbt]
\centering
\includegraphics[width=1\columnwidth]{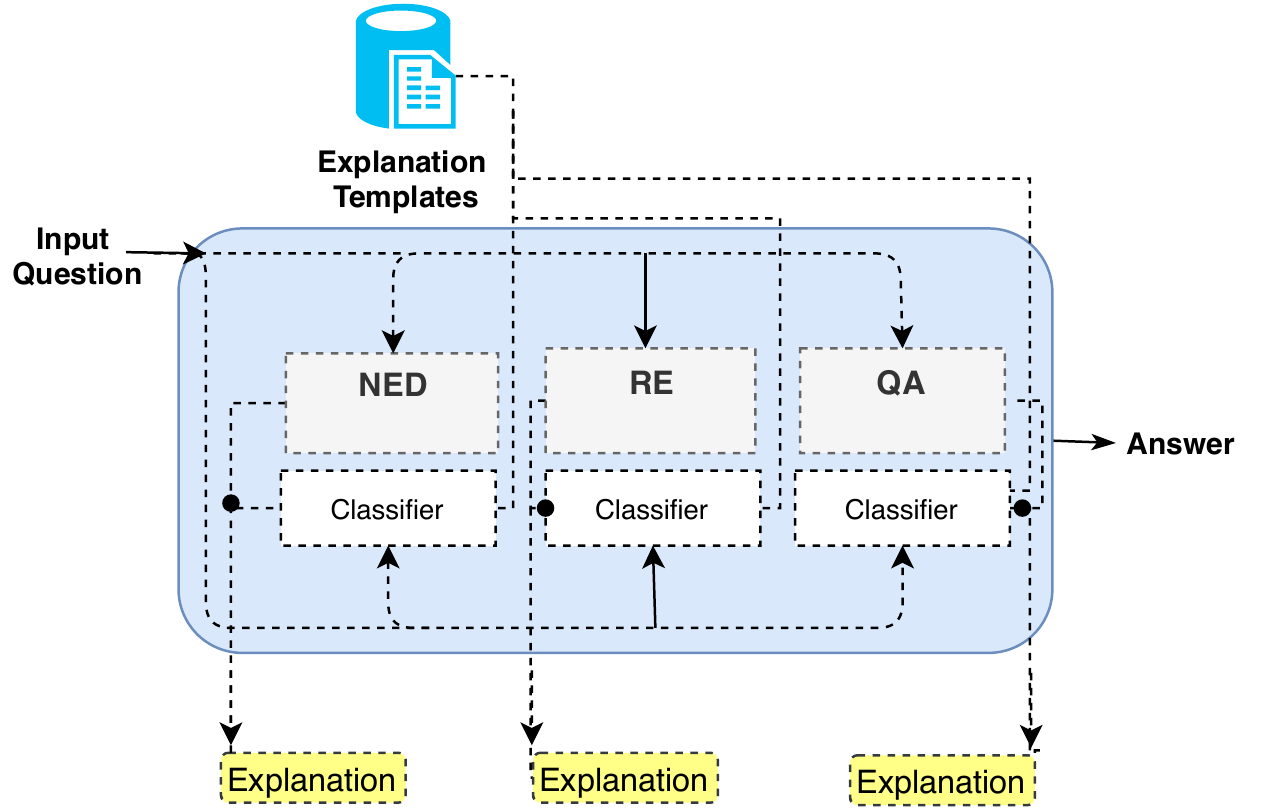}
\caption{This figure sketches a top overview of our approach. There is a classifier for each component, which predicts the output of the associated component. Also, there is a repository of the explanation templates. Thus, based on the prediction of the classifier and the actual output of the component, a suitable template is filtered.
For final explanation, the output of the component was incorporated into the template.}
\label{fig:architecture}
\end{figure}

\subsection{Methodology}
Figure \ref{fig:architecture} shows the architecture of our approach.
Initially, a pipeline for a QA system is built up; in our case we used Frankenstein platform \cite{www/SinghRBSLUVKP0V18,esws/SinghBRS18} to facilitate building up a pipeline. Please note, we do not aim to build a new QA system and reused an existing implementation. We extend the Frankenstein QA pipeline as illustrated in Figure \ref{fig:architecture}. 
We rely on the best performing pipeline reported in \cite{www/SinghRBSLUVKP0V18} over LC-QuAD dataset \cite{trivedi2017lc}.
In addition, we manually populated a repository of explanation templates. For example, all the required explanation templates for NED components are created for cases such as templates for wrong answers, when components produce no answer, and in the case of correct answers. Similarly, the templates for other tasks such as RE an QB were handcrafted. Please note that these templates are generic, thereby they do not depend on the employed component. For example, if we integrate another NED component rather than TagMe, there is no need to update the template repositories. 
In the next step, we trained classifiers based on the settings which will be presented in the next section.
Thus, when a new question arrives at the pipeline, in addition to running the pipeline to exploit the answer, our trained classifiers are also executed.
Then the predictions of the classifiers lead us to choose appropriate templates from the repositories. The filtered templates incorporate the output of the components to produce salient representations for NL explanations. 
The flow of the explanations is represented to the end user besides the final answer.

\textbf{Templates for Explanation}
To support our approach for explainable QA, we handcrafted 11 different templates for the explanation. We create placeholders in the predefined templates to verbalize the output of the QA components. Consider the explanation provided in Figure \ref{fig:example}. The original template for explaining the output of TagMe component is: \texttt{TagMe identifies the multiword \textbf{X} as the entity in the question. The entity is mapped to the DBpedia concept \textbf{dbr:W}.} The placeholders \textbf{X} and \textbf{dbr:W} are replaced accordingly for each question if a classifier selects this template in its prediction. 

\section{Experimental Study}
\label{sec:experiment}

We direct our experiment in response to our two research questions (\textit{i.e.,} RQ1 and RQ2) respectively. First, we pursue the following question \question{How effective is our approach for generating explanations?} This evaluation implies the demonstration of the success of our approach in generating proper explanations.
It quantitatively evaluates the effectiveness of our approach. 
On the contrary, the second discourse of the experiment is an HCI study in response to the question \question{How effective is the perception of the end user on our explanations?} This experiment qualitatively evaluates user perception based on the human factors introduced earlier (cf. Section 1).
In the following Subsections, we detail our experimental setups, achieved results, and insights over the outcomes of the evaluation. 

\subsection{Quantitative Evaluation}
This experiment is concerned with the question \question{How effective is our approach for generating explanations?}. We measure the effectiveness in terms of the preciseness of the explanations. 
Regarding the architecture of our approach, choosing the right explanation template depends on the prediction of the classifiers. 
If classifiers precisely predict a correct output for the underlying components, then consequently, the right templates will be chosen.
\begin{figure}[hpbt]
\centering
\includegraphics[width=\columnwidth]{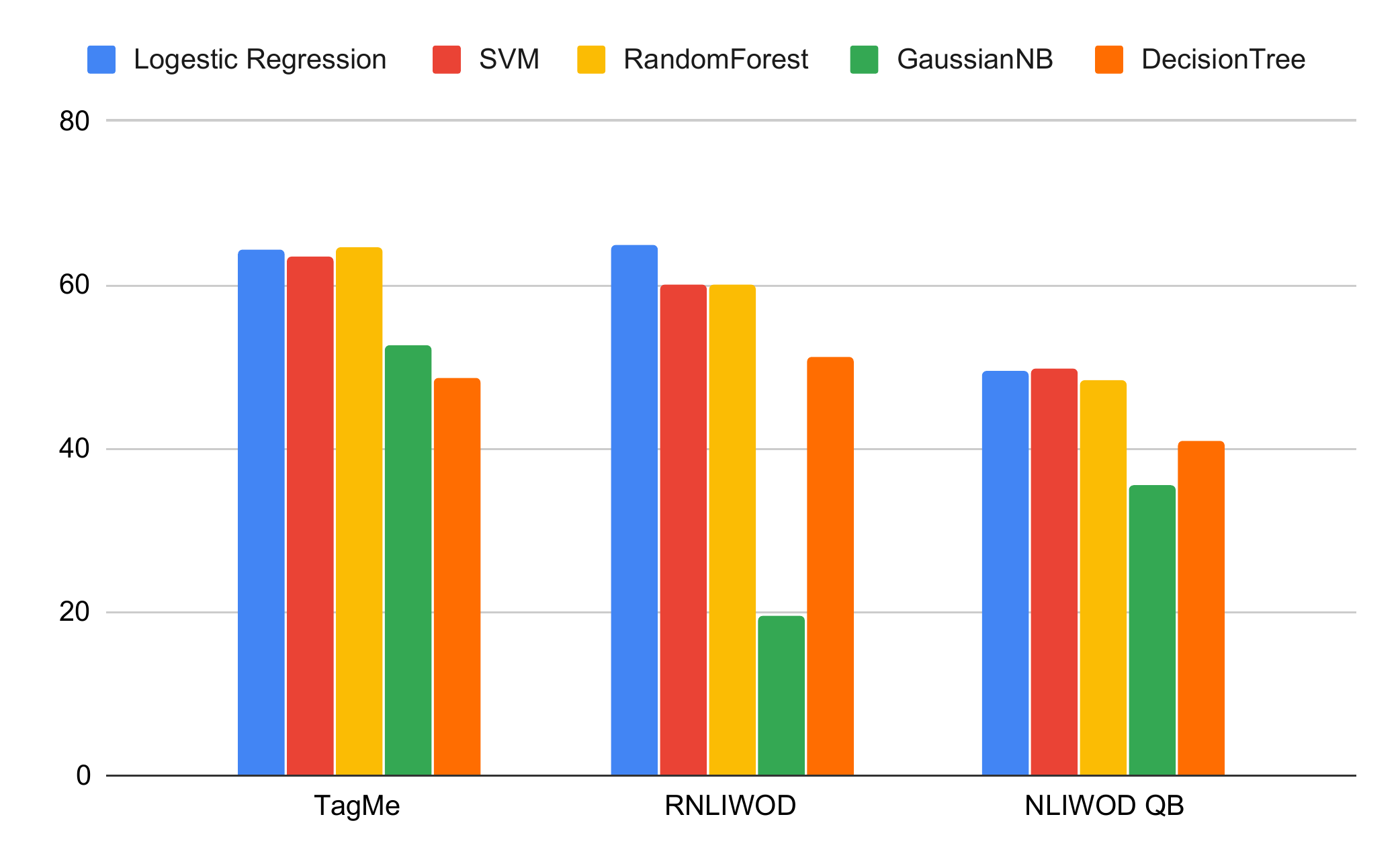}
\caption{This figure illustrates the accuracy of five classifiers per QA component: TagMe, RNLIWOD, and NLIWOD QB. Logistic Regression classifier performs best for all the components.}
\label{fig:chart}
\end{figure}

In other words, any flaw in the prediction leads to a wrong template. 
Thus, here we present the accuracy of our classifiers per component. We consider three generic classes, namely $O_c=\{Success, No Answer, Wrong Answer\}$ (cf. section \ref{sec:approach}) for the outputs of individual components. A benchmarking approach has been followed to choose best classifier per task. We employ five different classifiers (SVM, Logistic Regression, Random Forest, Gaussian NB, and Decision Tree) and calculated each classifier's accuracy per component. To train the classifiers per component, we require to create a single dataset. The sample-set in training is formed by considering questions of the LC-QuAD dataset. To get the concrete representation of each question, we extracted the following features: question length, headword(who, what, how), answer types (boolean, number, list), and POS tags. If a particular feature is present, we consider the value 1; if not, then the value of that feature is 0 while representing the question. The label set of the training datasets for a given component was set up by
measuring the micro F-Score of every given question for 3,253 questions from the LC-QuAD dataset. The F-score per question is calculated by adopting the same methodology proposed by \cite{www/SinghRBSLUVKP0V18}. We rely on 3,253 questions out of 5,000 questions of the LC-QuAD dataset because the gold standard SPARQL queries of the remaining 1,747 questions do not return any answer from DBpedia endpoint (also reported by \cite{azmy2018farewell}). The classifier predicts if a component can answer the question or not, and trained using features extracted from the natural language questions against the F score per question. During the training
phase, each classifier was tuned with a range of regularization on the dataset. We used the cross-validation approach with 10
folds on the LC-QuAD dataset.
We employ a QA pipeline containing TagMe \cite{DBLP:conf/cikm/FerraginaS10} for entity disambiguation, RNLIWOD\footnote{Component is similar to Relation Linker of \url{https://github.com/dice-group/NLIWOD}} for relation linking, and NLIWOD QB\footnote{Component is based on \url{https://github.com/dice-group/NLIWOD} and \cite{DBLP:conf/www/UngerBLNGC12}.} for SPARQL query builder. Figure \ref{fig:chart} reports the accuracy of five classifiers (average of all classes). Furthermore, Table \ref{tab:multi-classClassification} reports the accuracy of the best classifier (Logistic Regression in our case) for each component.

\begin{table}[!h]
\centering
\begin{scriptsize}
\begin{tabular}{ll}
\toprule
\rowcolor{Gray}  \textbf{Component}          & \textbf{Accuracy}  \\ \midrule
\textbf{TagMe} & 0.64  \\ \hline
\textbf{RNLIWOD}  &   0.60  \\ \hline
\textbf{NLIWOD WB}    &  0.49 \\ \hline
%\\ \bottomrule
\end{tabular}
\end{scriptsize}
\caption{\bf Accuracy of our multi-class classifier for predicting type of explanation for each component.}
\label{tab:multi-classClassification}
\end{table}

\paragraph{Observations.} We observe that the logistic regression classifier performs best for predicting the output of components.
%choosing the right explanation template across tasks. 
However, the accuracy of the classifier is low as depicted in the Table \ref{tab:multi-classClassification}. \cite{www/SinghRBSLUVKP0V18} report accuracy of \textit{binary classifiers} for TagMe, RNLIWOD, and NLIWOD QB as 0.75, 0.72, and 0.65 respectively. When we train \textit{multi-class classifiers} (\textit{i.e.,} three classes) on the same dataset, we observe a drop in the accuracy. The main reason for the low performance of the classifiers is the low component accuracy (c.f. Table \ref{tab:motivating})

\subsection{User Perception Evaluation}
In the second experiment, we pursue the following research question: \question{How is the perception of end user about explanations along the human factor dimensions?} To respond to this question, we conduct the following experiment:\\
\textbf{Experimental Setup}: We perform a user study to evaluate how the explanations impact user perception.
We aim at understanding user's feedback on the following four parameters inspired by \cite{DBLP:conf/iui/EhsanTCHR19,exai}: 1) \texttt{Adequate Justification}:  Does a user feel the answer to a particular question is justified or provided with the reasoning behind inferences of the answer? 2) \texttt{Education}: Does the user feel educated about the answer generation process so that she may better understand the strengths and limitations of the QA system? 3) \texttt{User involvement}: Does the user feel involved in allowing the user to add her knowledge and inference skills to the complete decision process? 4) \texttt{Acceptance}: Do explanations lead to a greater acceptance of the QA system in future interactions?
With respect to the above criteria, we created an online survey to collect user feedback.
The survey embraces random ten questions from our underlying dataset from a variety of answer types such as questions with the correct answer, incorrect answer, no answer (for which classifiers predict correct templates). 
The first part of the survey displays the questions to the user without any explanation. In the second part, the same ten questions, coupled with the explanations generated by our approach, are displayed to the user. The participants of the survey are asked to rate each representation of question/answer based on the four human factor dimensions (i.e., acceptance, justification, user involvement, and education). 
The rating scale is based on the Likert scale, which allows the participants to express how much they agree or disagree with a given statement (1:strongly disagree -- 5:strongly agree).
We circulated the survey to several channels of the co-authors' network, such as a graduate class of Semantic Web course, research groups in the USA and Europe, along with scientific mailing lists. Collectively we received responses from 80 participants. Please note, the number of participants is at par with the other explainable studies such as \cite{DBLP:conf/iui/EhsanTCHR19}.

%All the five questions required different explanations including successful answer, failure due to wrong entity linking, failure due to wrong predicate linking, failure due to no answer extraction, and failure due to wrong formulation of the SPARQL Query. 
%We have two settings: 1) five questions from the dataset and its answers generated by the QA pipeline are displayed to the user without explanation 2) same five questions coupled with the explanation generated by the prediction of the classifiers in previous experiments are provided to the user. Users are asked to rate the four parameters on the scale of 1--5 (scale point five is strongly agree, scale point one is strongly disagree). All the five questions required different explanations including successful answer, failure due to wrong entity linking, failure due to wrong predicate linking, failure due to no answer extraction, and failure due to wrong formulation of the SPARQL Query. 

\begin{figure}[hpbt]
\centering
    \includegraphics[width=\columnwidth]{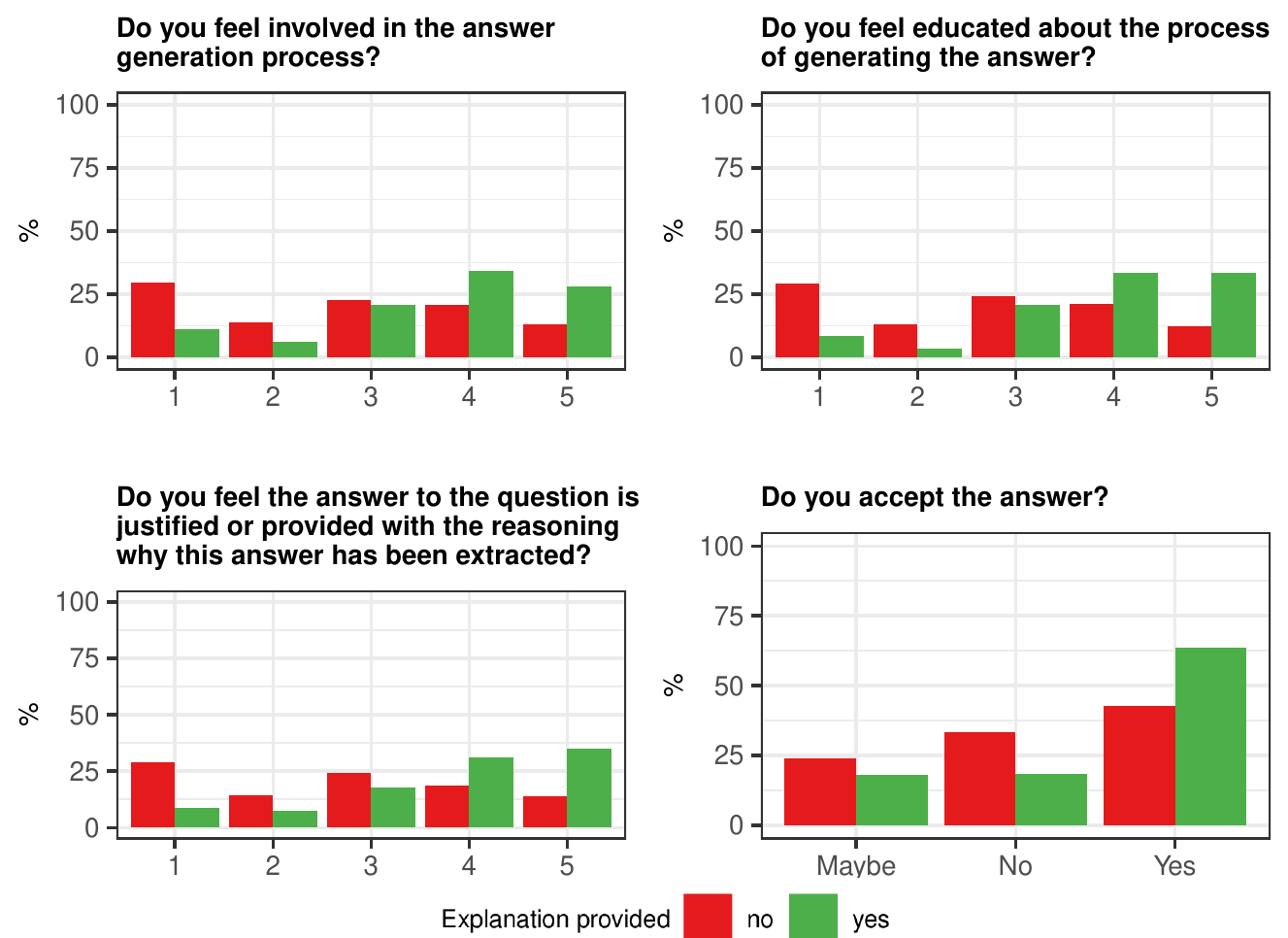}
    \caption{User perception Evaluation. The figure illustrates the comparative analysis of providing with and without explanation to the user. We consider the mean of all the responses. X-axis depicts the Likert scale (1 is strongly disagree, 5 is strongly agree). A clear trend in user responses shows that across all four parameters, there are many answers towards disagreement or neutral when no explanation is provided. In the case of explanation, users feel involved, and responses are shifted towards the agreement. Furthermore, users show more trust in the acceptance of the answer when provided with an explanation.}
     \label{fig:user}
    \end{figure}

\paragraph{Results and Insights.}
Figure \ref{fig:user} summarizes the ratings of our user study. We evaluate the user responses based on the four human factor dimensions: \texttt{Adequate Justification}, \texttt{Education}, \texttt{User involvement}, and \texttt{Acceptance}. The summary of ratings for each dimension was captured in one individual chart. 
The green bars show the feedback over questions with provided explanations, and on the contrary, red bars are aggregated over the question with no explanation.
The x-axis shows the Likert scale. The Y-axis is the distribution of users over the Likert scale for each class independently- with explanation and without explanation. 
Overall it shows a positive trend towards the agreement with the following facts; the provided explanations helped users to understand the underlying process better, justify a particular answer, involve the user in the complete process, and increase the acceptability of the answers. The green bars are larger in positive ratings, such as strongly agree.

%%%%%%%%%%%%%%%%%%%%%%%%%%%%%%%%%%%%%%%
\section{Discussion}
\label{sec:discussion}

In this paper, we focus on the challenge of explainable QA systems.
We mainly target systems that consume data from the KGs.
These systems receive a natural language question and then transform that to a formal query.
Our primary aim is to take the initial steps to break down the full black-box QA systems.
Thus, we reuse an existing QA pipeline systems since it already decompose the prominent tasks of the QA systems and then integrate individual implementations for each QA task. We based our approach and associated evaluation on the hypothesis that every component integrated into the pipeline should explain the output. It will educate and involve non-expert users and trigger them to trust and accept the system. Our findings in Section \ref{sec:experiment} support our hypothesis both on quantitative and qualitative evaluation.
The limitation of our approach is that it heavily relies on the performance of the components. In the case of having low performing components, the accuracy of the classifiers is also downgraded.
Although, on the one hand, this approach is shallow, one the other hand it avoids exposing the user to overwhelming details of the internal functionalities by showing succinct and user-friendly explanations. \cite{hoffman2017explaining} noted that for improving the usability of XAI systems, it is essential to combine theories from social science and cognitive decision making to validate the intuition of what constitutes a "good explanation." Our work in this paper is limited to predefined template based explanations, and does not consider this aspect. Also, our work does not focus on the explainability of the behavior of the employed classifier, and the explanations only justify the final output of components. 

%currently the classifiers are limited to predict one dimension of components. Still there are other aspects which have not been involved such cases that the answer is partically correct, cases that.
%We proposed an approach which provides explanations for pipeline
%1. Reason to choose pipeline to demonstrate shallow insights whats happening insight
%2. we reused existing implementation of a framework, and do not propose new QA system. Instead, we show how the user perception improves if provided with simple explanations. 
%3. we contribute to the methodology of semi- AI based method for generating explanation. 
%4. Our approach heavily relies on the performance of the components. 
%5. If we also want to integrate a new component, the training of classifier is from scratch. 
%Classifiers are dummy, they don't different -- 
%completeness of answer
%personalize explanation or contextualized explanation

\section{Conclusion and Future Direction}
In this paper, we proposed an approach that is automatic and supervised for generating explanations for a QA pipeline. Albeit simple, our approach intuitively expressive for the end user. 
This approach requires to train a classifier for every integrated component, which is costly in case the components are updated (new release) or replaced by a latest outperforming component.
Our proposed approach induced in a QA pipeline of a modular framework is the first attempt for explainable QA systems over KGs. It paves the way for future contributions in developing explainable QA systems over KGs.
Still, there are numerous rooms in this area that require the attention of the research community -- for example, explanations regarding the quality of data, or metadata, or credibility of data publishers. Furthermore, recent attempts have been made to provide explanations of machine learning models \cite{guo2018explaining}. However, the inclusion of the explanations in neural approaches for question answering (such as in \cite{www/LukovnikovFLA17}) is still an open research question, and we plan to extend our work in this direction. 
The concerning domain of the system is also influential in explanations. for example, biomedical or marketing domains require various levels of details of explanations.
In general, all of these concerns affect the acceptance and trust of the QA system by the end user.
Our ambitious vision is to provide personalized and contextualized explanations, where the user feels more involved and educated.

\label{sec:conclusion}